\documentclass[letterpaper, 10 pt, conference]{ieeeconf}  % Comment this line out if you need a4paper

\IEEEoverridecommandlockouts                              % This command is only needed if 
\usepackage[usenames, dvipsnames]{color}
\usepackage[dvipsnames]{xcolor}
\usepackage{amsmath}
\usepackage{mathtools}
\usepackage{amsfonts}
\usepackage{amssymb, bm}
\usepackage{tikz}
\usetikzlibrary{positioning,shapes,arrows}
\usepackage{graphicx}
\usepackage[hyphens]{url}
\usepackage{hyperref}
\usepackage[caption=false]{subfig}
\usepackage{cite}
\DeclareMathOperator*{\argmin}{argmin}
\title{\bf Learning of Coordination Policies for Robotic Swarms}

\author{Qiyang Li, Xintong Du, Yizhou Huang, Quinlan Sykora, and Angela P. Schoellig\thanks{The  authors  are  with  the  Dynamic  Systems  Lab  (\href{www.dynsyslab.org}{www.dynsyslab.org}), Institute for Aerospace Studies, University of Toronto, Canada. Email: qiyang.li@mail.utoronto.ca, xintong.du@mail.utoronto.ca, philipyizhou.huang@mail.utoronto.ca, quinlan.sykora@mail.utoronto.ca, schoellig@utias.utoronto.ca.}}

\begin{document}
\maketitle

\begin{abstract}
Inspired by biological swarms, robotic swarms are envisioned to solve real-world problems that are difficult for individual agents. Biological swarms can achieve collective intelligence based on local interactions and simple rules; however, designing effective distributed policies for large-scale robotic swarms to achieve a global objective can be challenging. Although it is often possible to design an optimal centralized strategy for smaller numbers of agents, those methods can fail as the number of agents increases. Motivated by the growing success of machine learning, we develop a deep learning approach that learns distributed coordination policies from centralized policies. In contrast to traditional distributed control approaches, which are usually based on human-designed policies for relatively simple tasks, this learning-based approach can be adapted to more difficult tasks. We demonstrate the efficacy of our proposed approach on two different tasks, the well-known rendezvous problem and a more difficult particle assignment problem. For the latter, no known distributed policy exists. From extensive simulations, it is shown that the performance of the learned coordination policies is comparable to the centralized policies, surpassing state-of-the-art distributed policies. Thereby, our proposed approach provides a promising alternative for real-world coordination problems that would be otherwise computationally expensive to solve or intangible to explore.
%for the rendezvous problem with robots having limited visibility. %Our proposed approach, based on learning from centralized policies, provides a promising alternative for real-world coordination problems that would be otherwise computationally expensive to solve or intangible to explore.
%With potentially similar performance compared to centralized policy, our proposed approach provides a promising alternative for real-world coordination problems that would be otherwise computationally expensive to solve or intangible to explore.
\end{abstract}

\section{Introduction}
Biological swarms can act in coordination to perform tasks far beyond the capabilities of individuals~\cite{bonabeau1999swarm}.
% through the principle of self-organization: ant swarms can intelligently forage for food; honey bee swarms can build combs.
In the absence of a centralized control mechanism and global observation, swarm intelligence emerges from the local behaviour of individual agents governed by simple, unified rules~\cite{camazine2003self}. The distributed structure of swarm systems makes them less vulnerable to individual failures. When this robust nature is well-realized in robotic systems, robotic swarms can then be relied upon to solve complex real-world problems such as search and rescue, object transportation, and Mars exploration, where centralized control can be extremely costly or may be impossible~\cite{murphy2008search,zhu2013return,campo2006negotiation}. 

%Complex real-world problems such as search and rescue, object transportation and Mars exploration require a robust robotic system with a large number of agents, where centralized control can be extremely costly or even unavailable
%This is where robotic swarms can excel with their highly distributed, robust control structures. % botic tasks  When swarm intelligence is applied to large number of agents, there can be many real-world applications, such as search and rescue, object transportation and especially Mars exploration, where no centralized algorithm is available \cite{zhu2013return,campo2006negotiation}.
% computaional cost for centralized control/distributed control
%We believe that robot swarms can also be designed such that desired collective behaviors emerge from local interactions among agents and local observations from the environment. %

\begin{figure}[t]
    \centering
    \vspace{0.8em}
    \includegraphics[width=8.5cm]{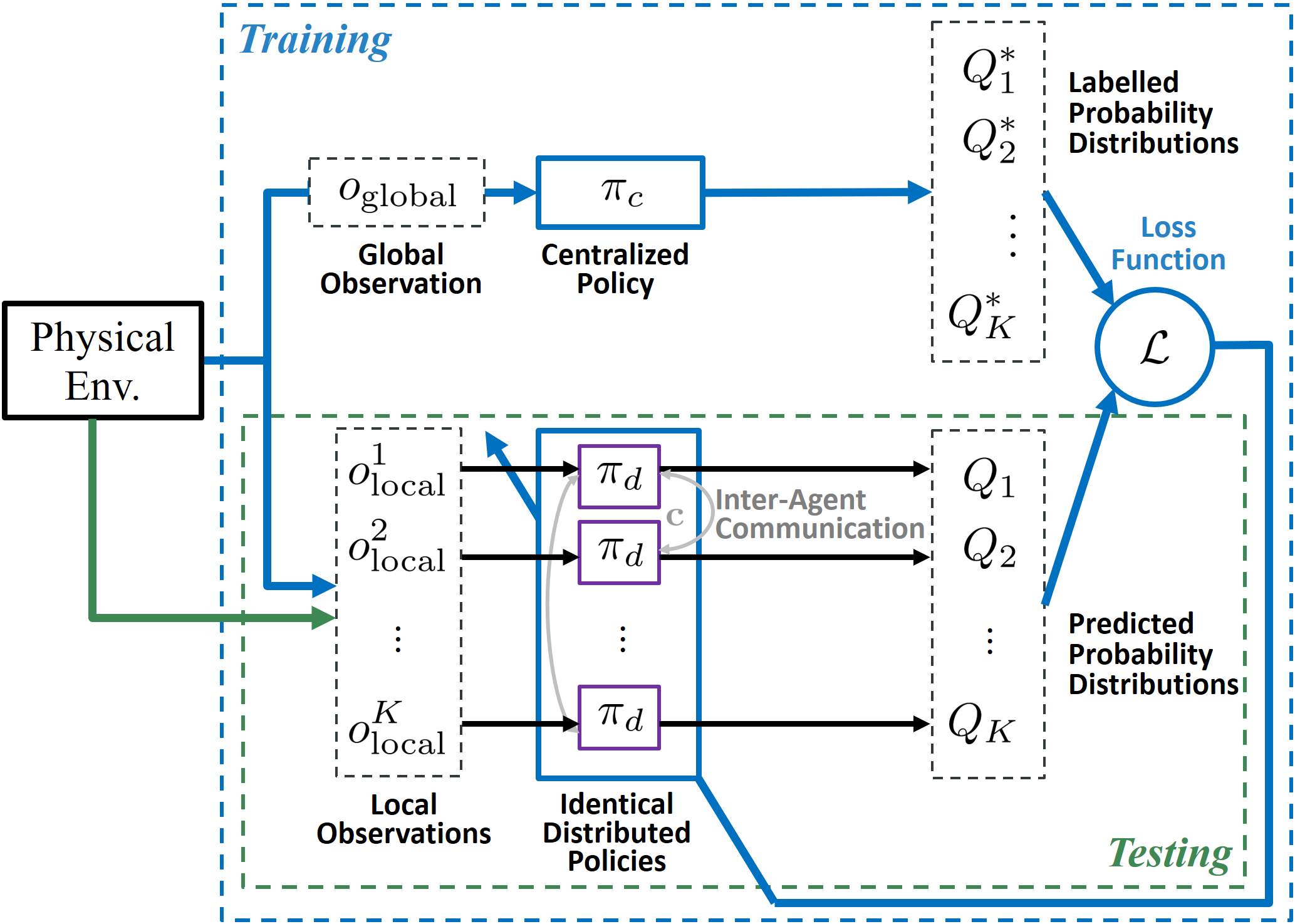}
    \caption{The proposed approach learns distributed control and communication strategies from well-designed centralized policies for $K$ identical agents.}
    \label{fig:first}
    \vspace{-1.5em}
\end{figure}

To explore this domain, this work focuses on studying distributed robotic systems with the following properties:
\begin{itemize}
    \item All agents are identical (i.e., homogeneous).
    \item Each agent has only local interactions and observations.
    \item Each agent follows an identical distributed policy with no presence of a leader.
\end{itemize}
%These properties bring several advantages: robustness, flexibility and relatively simple, similar individual agents. They are robust, as the system is less susceptible to failures of individual agents and do not require a centralized control unit with sophisticated computational capabilities. Large numbers of simpler, mono-type individual agents compared to a single complex system are less prone to noise due to the multiplicity of sensing\cite{csahin2004swarm}. %These advantages motivate our research on homogeneous, distributed robotics systems.% - the construction of inter-agent communication protocols. %to relay local action and observations to others.
To coordinate individual agents towards achieving a mutual objective, a key challenge is to relay local actions and observations from agent to agent. A distributed policy, therefore, contains two main components: an action policy that determines what action the agent performs given its inputs, and a communication protocol that defines how agents communicate with each other.
To achieve system-level coordination, classical methods rely on human-designed policies, in which communication protocols are usually predefined~\cite{cao2013overview}. This manual design process can be especially difficult for complex coordination tasks with large swarms.
%swarm robotic tasks due to the inherent complexity of swarm systems.
Recently, learning-based approaches, such as deep reinforcement learning, have been able to successfully learn communication protocols in coordination tasks~\cite{kasai2008learning, lazaridou2016multi, mordatch2017emergence, lowe2017multi, giles2002learning, foerster2016learning, Foerster2016, sukhbaatar2016learning, das2017learning}. The biggest drawback of these approaches is the tremendous amount of data and computational resources required for training. %Even with a small number of agents on simple coordination tasks, training processes can take days or weeks.

In this paper, we propose a learning-based approach that utilizes pre-designed centralized policies to train the distributed policies. Each agent follows an identical distributed policy, which \textit{(i)} interprets the agent's observation and communication information received, \textit{(ii)} determines the action of the agent, and \textit{(iii)} generates the communication information to be broadcasted to the neighbouring agents. This distributed policy is modelled by a differentiable deep neural network (DNN) called \textit{distributed policy network}, where its inputs and outputs are represented as fixed-sized vectors. %The vectors of communication information sending from one distributed policy network to another distributed policy network are the communication channels. 
The distributed policy network serves as a part of a larger neural network that maps from all agents' multi-step observations to their multi-step actions (see Fig.~\ref{fig:giant_nn}). We refer to this augmented neural network as the multi-step, multi-agent neural network (MSMANN). Since communication vectors sent from agent to agent are hidden states of the MSMANN, we can perform backpropagation on the MSMANN to capture the communication protocols in addition to the action policies, which completes the distributed policy we aim to learn. Since the input and output vectors for the communication information all have fixed sizes, the distributed policy network must learn how to aggregate information for effective communication. This also enables us to control the communication flow and analyze the communication information sent from agent to agent. This can be regarded as an analogy to controlling the size of the representation layer of an auto-encoder and analyzing the learned representation. %the differentiable communication channels across multiple time steps to capture the communication protocols in addition to the action policies, which complete the distributed policies.

Our learning approach can be applied to a variety of distributed robotic tasks, provided that a well-designed centralized policy is available. %We validate our approach on two distributed robotic tasks: rendezvous task with limited visibility and particle assignment. 
On a rendezvous task with limited visibility, this approach consistently outperforms the state-of-the-art distributed control law for systems consisting of different numbers of agents. %The feasibility of our approach on another distributed robotic task we define  illustrates the potential of applying this approach in robotic tasks with no existing distributed solutions. 
The generalizability of our proposed approach to other robotic tasks is demonstrated in a task for which no distributed solution exists. Our approach provides a tangible alternative for complex robotic tasks for which centralized policies can be designed. %\textbf{[motivate a bit more about centralized policy over distributed polciy]} 
Moreover, we analyze the learned communication protocols and provide insights on their meaning. This analysis of learned  coordination policies could potentially benefit the manual design of communication protocols for complex robotic swarm systems.

In the following sections, we begin discussions with a brief overview of related literature in Section \ref{sec:relatedwork}. Upon defining the control framework and the problem statement in Section \ref{sec:framework} and \ref{sec:problem}, our learning approach is discussed in Section \ref{sec:methodology}. The simulation setup and the corresponding results are presented in Section \ref{sec:simulation_setup} and \ref{sec:simulation_result}.

\section{Related Work}
\label{sec:relatedwork}
In the distributed control literature, most work focused on the manual design and analysis of distributed control laws for simple robotic tasks~\cite{cao2013overview}. For example, \cite{francis2016flocking} and \cite{bullo2009distributed} both study rendezvous and flocking tasks and analyze the performance of various distributed control laws. However, these studies are based on simplified robot dynamics (e.g., single or double integrator). Adapting these approaches to more complicated distributed robotic tasks can be challenging.
%However, these distributed control laws were usually designed manually for tasks with simple specifications but not for more difficult scenarios. The reason is that when the distributed robotic tasks become more complicated, the manual design of control algorithm can be very challenging.

With little human expertise required, learning-based approaches provide alternatives for challenging distributed robotic tasks. The majority of these works focus on only learning the action policy, and assume pre-defined communication protocols or no communication among the agents at all~\cite{chen2017decentralized, zhang2013coordinating, maravall2013coordination}. These approaches lack the flexibility to adapt the communication information, thus, limiting the capability of the distributed robotic system. Our approach learns both the control policy and the communication protocols among the agents, %This eliminates the need of designing distributed policies while 
allowing the emergence of more effective inter-agent communication.

Some recent work has explored learning communication protocols through reinforcement learning. Specifically, \cite{das2017learning} and \cite{kasai2008learning} perform independent learning where each agent learns solely from its local observations. % perform tabular Q-learning on 2 agents. Update alternatively.
Others assume that centralized learning can be performed~\cite{lowe2017multi,mordatch2017emergence,lazaridou2016multi} and \cite{sukhbaatar2016learning}.
%Others assume that centralized learning can be performed. In \cite{lowe2017multi}, centralized critics are used to perform actor critic reinforcement learning. In \cite{sukhbaatar2016learning, lazaridou2016multi, mordatch2017emergence}, policy gradient is performed across multiple agents in a centralized manner. In \cite{Foerster2016, foerster2016learning}, Q-learning is also performed in a centralized manner. 
In particular, \cite{sukhbaatar2016learning}, \cite{mordatch2017emergence} and \cite{Foerster2016} assume that the communication channel can be differentiated during backpropagation. While \cite{sukhbaatar2016learning} employs multiple communication cycles at each time step, \cite{mordatch2017emergence} and \cite{Foerster2016} are similar to our control framework, which assumes that the communication emitted can only be received at the next time step. 

Different from all the reinforcement learning approaches above, our approach makes use of pre-designed centralized policies and learns communication protocols by imitating the behaviour of the centralized policies. Compared to the reinforcement learning approaches, our approach can learn communication and action policies more efficiently using the guidance of the centralized policies. This approach is inspired by \cite{sukhbaatar2016learning}, which performs supervised learning on a simple lever-pulling task using
multiple communication cycles at each time step. We perform learning in the distributed robotic
domain using a different framework, which assumes a single
communication cycle at each time step.

\section{Distributed Control Framework}
\label{sec:framework}
Consider a group of $K$ homogeneous agents that are to complete a task $T$ with a dynamic connectivity network defined by a bidirectional graph $G(t) = (V, E(t))$, where $t = \{1, 2, \cdots, L\}$ is the discrete time index. Each $v_i \in V$, $V = \{v_1, v_2, \cdots, v_K\}$, represents an agent and each $e(v_i, v_j) \in E(t)$ represents a connection between agent $v_i$ and agent $v_j$. 

We define $N_i(t) = \{v_j | e(v_i, v_j) \in E(t)\}$ to be the set of all the neighbours of agent $v_i$. For each neighbour $v_j \in N_i(t)$, we can define $\mathbf{c}_{i  j}(t)$ to be the communication information sent from agent $v_i$ to agent $v_j$ at time $t$. In this work, we define $\mathbf{c}_{i  j}(t) \in \mathbb{R}^{n}$ as a communication vector, where $n$ is the size of the communication information. 

For convenience, we also define
%\begin{equation}
%    c^i_{\text{in}}(t) = \begin{cases} 
%      \{\mathbf{c}_{j  i}(t-1) | v_j \in N_i(t-1)\} & t > 1  \\
%      \emptyset & t = 1 
%   \end{cases}
%\frac{1}{N(v_i)}\sum_{v_j \in N(v_i)}^{} c_{v_j, v_i}
%\end{equation}
\begin{equation}
    c^i_{\text{in}}(t) = \{ \mathbf{c}_{j  i}(t-1) | v_j \in N_i(t-1)\}, t > 1,
\end{equation}
to be the communication inflow to $v_i$, with $c^i_{\text{in}}(1) = \emptyset$, and 
\begin{equation}
c^i_{\text{out}}(t) = \{ \mathbf{c}_{i  j}(t) | v_j\in N_i(t)\}
\end{equation}
%\begin{equation}
%    c^i_{\text{out}}(t) = \{ \mathbf{c}_{i  j}(t) | v_j\in N_i(t)\} %\frac{1}{N(v_i)}\sum_{v_j \in N(v_i)}^{} c_{v_i, v_j}
%\end{equation}
to be the communication outflow from $v_i$. This means that the communication information sent at the previous time step $(t-1)$ is received at the current time step $t$. When $v_i$ and $v_j$ are the same agent, the communication vector $\mathbf{c}_{ij}$ is a special case, which allows the agent to leave information to itself. In this work, we allow this self-talking behaviour.

\begin{figure}[t]
    \centering
    \includegraphics[width=0.45\textwidth]{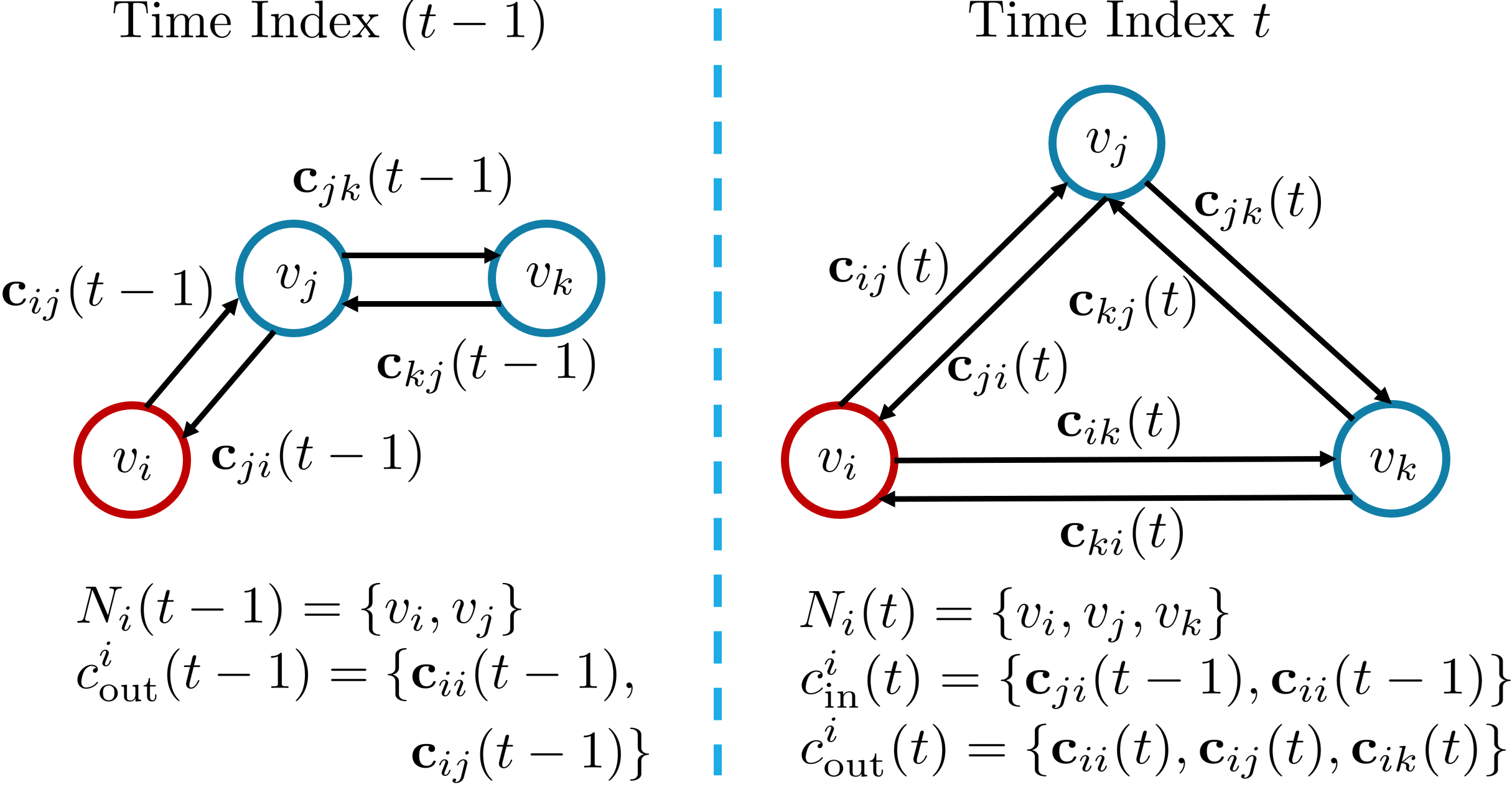}
    \caption{The communication framework with three agents at two consecutive time steps. The connectivity network is indicated by the arrows among the agents. We show the neighbours and the communication associated with $v_i$ for each time step. At time index $(t-1)$, agent $v_i$ can only communicate to agent $v_j$ and send information to itself. At time index $t$, agent $v_i$ receives the information sent by $v_j$ from the previous time step and sends out information to its neighbours, $v_j$ and $v_k$, and to itself.}
    \label{fig:example}
\end{figure}

At each time step, agent $v_i$ obtains an observation information $o^i_{\text{local}}$ and receives information $c^i_{\text{in}}$. The agent then sends information, $c^i_{\text{out}}$, to its corresponding neighbours $N_{i}$, and performs an action $a_i \in \mathcal{A}$, where $\mathcal{A}$ is the unified action space of all agents (see Fig.~\ref{fig:example}). 

\begin{figure}[t]
    \centering
    \includegraphics[width=0.45\textwidth]{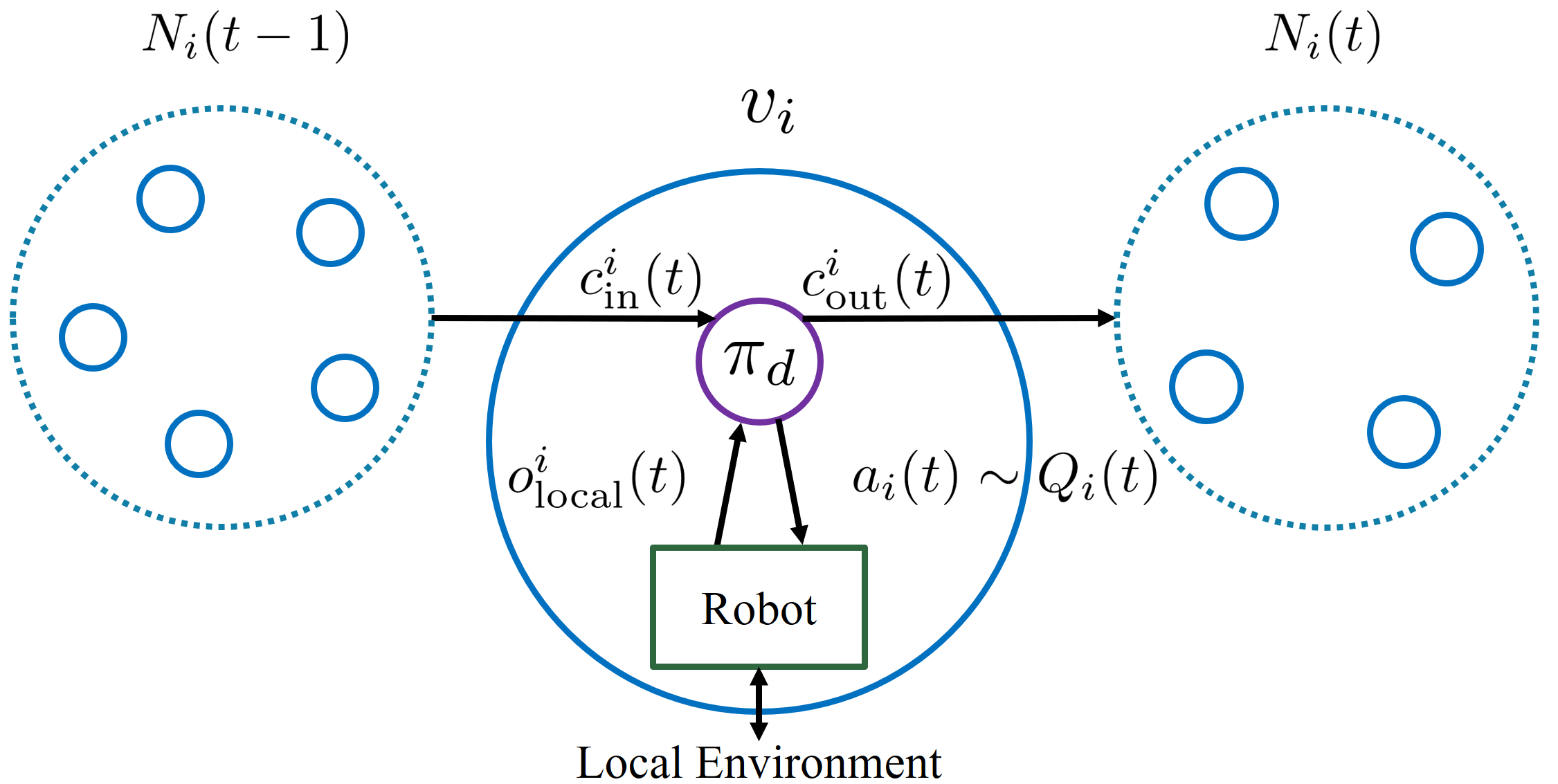}
    \caption{Overall picture of the homogeneous, distributed control framework. The agent receives the communication inflow $c^i_{\text{in}}(t)$ from its neighbours $N_i(t-1)$ from the previous time step and observes the surrounding $o^i_{\text{local}}(t)$. Given the inputs, the policy then provides the communication outflow $c^i_{\text{out}}(t)$ and the action probability distribution $Q_i(t)$. The communication outflow is sent to the agent's neighbours $N_i(t)$. The action of the agent $a_i(t)$ is sampled based on the probability distribution $Q_i(t)$.}
    \label{fig:overall}
    \vspace{-1em}
\end{figure}
\begin{figure}[t]
    \centering
    \includegraphics[width=0.35\textwidth]{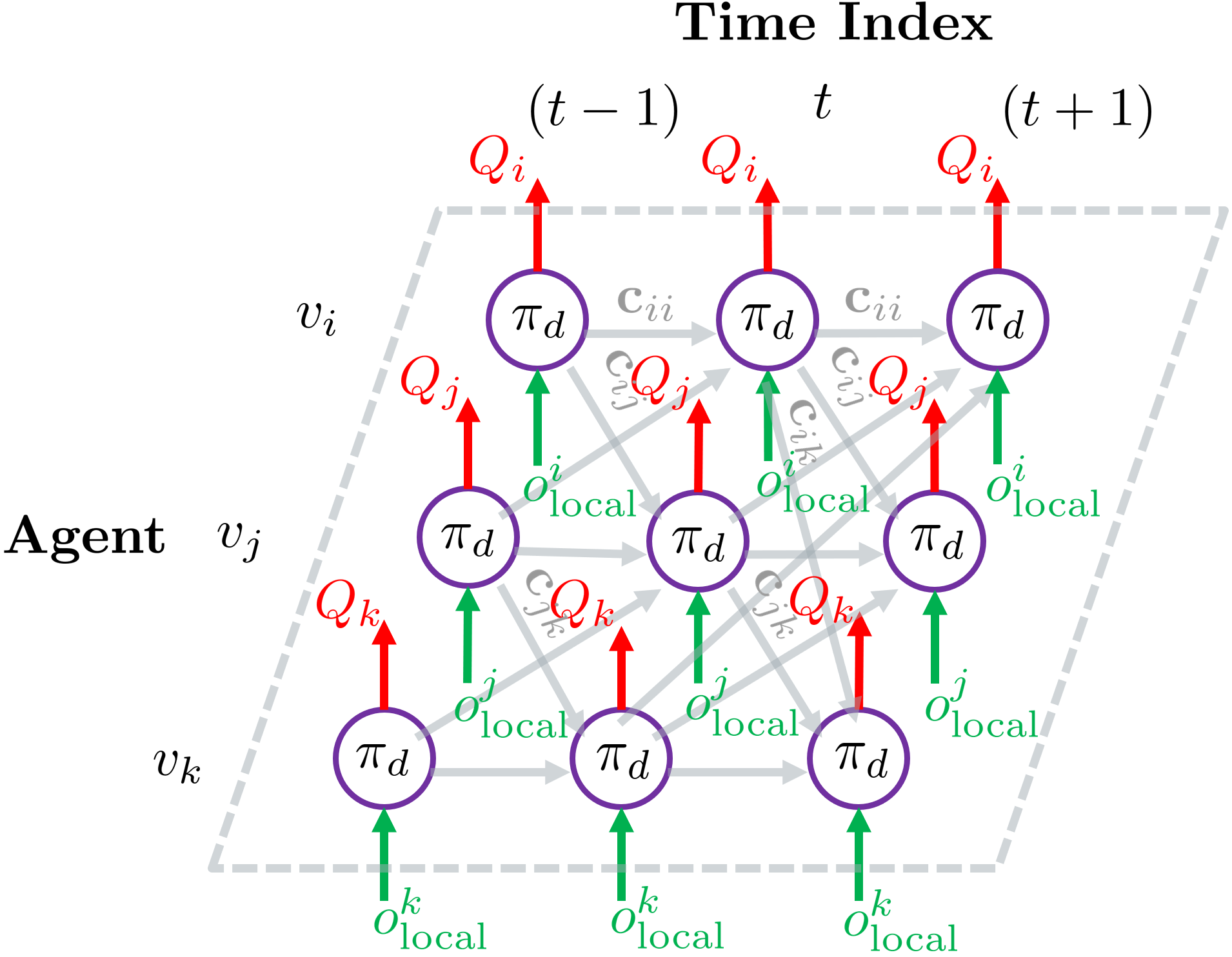}
    \caption{Multi-step, multi-agent neural network (MSMANN) consisting of identical components $\pi_d$. This is an example of the MSMANN for the three-agent example in Fig.~\ref{fig:example}. The grey arrows represent the communication vectors, which are also hidden states of the MSMANN; the green arrows represent the inputs (local observation $o_{\text{local}}$) of the neural network and the red arrows represent the outputs (action probability distribution $Q$) of the neural network.}
    \label{fig:giant_nn}
    \vspace{-1em}
\end{figure}

We define $\pi_d$ as a unified, distributed policy for all agents with the following equations holding true for all $v_i \in V$:
\begin{align}
    \pi_d(o^i_{\text{local}}(t), c^i_{\text{in}}(t)) &= (Q_i(t), c^i_{\text{out}}(t)) \\
    a_i(t) &\sim Q_i(t), 
\end{align}
where $Q_i(t)$ is the probability distribution over the action space $\mathcal{A}$ used by $v_i$ at time $t$ (see Fig.~\ref{fig:overall}). 

\section{Problem Statement}
\label{sec:problem}
Consider $K$ agents with the homogeneous, distributed control framework described in Section \ref{sec:framework}. The problem targeted by this work is to learn a mapping from communication inflow $c_{\text{in}}$ and local observations $o_{\text{local}}$ to communication outflow $c_{\text{out}}$ and action probability distribution $Q$. This mapping is referred to as the distributed policy $\pi_d$ modelled by a DNN, which interprets the communication inflow and the local observations, and infers the communication outflow and the action probability distribution. The goal is to find a policy that optimizes the performance of the target task $T$. During the learning process, robot actions and observations of a centralized strategy are available (Fig.~\ref{fig:first}). %In parallel, the local observations and actions of all agents are recorded for training. During testing, only local observation and communication is available.

\section{Methodology}
\label{sec:methodology}
\subsection{Neighbour Discretization}
The amount of communication inflow and outflow for each agent is changing dynamically as the number of its neighbours changes. Since learning a model that has variable input and output dimensions can be challenging, we perform a neighbour discretization to provide constant input/output dimensions. In this process, each agent's neighbours are partitioned into $P$ groups based on a discretization rule $\mathcal{D}$:%which maps from the agent's input (observation and communication inflow) to the grouping of its neighbours: %\footnote{$P$ is not changing with time $t$.}
% \begin{equation}
%     \mathcal{D}(o^i_{\text{local}}, c^i_{\text{in}}) = \{\mathcal{P}^i_1, \mathcal{P}^i_2, \cdots, \mathcal{P}^i_P \}, \forall v_i \in V,
% \end{equation}
\begin{equation}
    \mathcal{D}(o^i_{\text{local}}): N_i \rightarrow \{\mathcal{P}^i_1, \mathcal{P}^i_2, \cdots, \mathcal{P}^i_P \}, \forall v_i \in V,
\end{equation}
% where
% \begin{align}
%     &\mathcal{P}^i_p \subseteq N_i, \forall p \in \{1, 2, \cdots, P\}, \\
%     &\mathcal{P}^i_1 \cup \mathcal{P}^i_2 \cup \cdots \cup \mathcal{P}^i_P = N_i,  \\
%     &\mathcal{P}^i_p \cap \mathcal{P}^i_q = \emptyset,  \forall p, q \in \{1, 2, \cdots, P\}, p \neq q.
% \end{align}
where each $\mathcal{P}^i_p \subseteq N_i$, $p \in \{1, 2, \cdots, P\}$, represents a subset of neighbouring nodes and $o^i_{\text{local}}$ may be used to obtain the discretization. To apply this discretization, we process the communication inflow and average the communication vectors within each group of agents. We also restrict the communication outflow such that the communication vectors sent to the agents in the same group are identical. To formulate this, we define the communication inflow $\mathbf{c}^{i, \mathcal{D}}_{\text{in}}$ and outflow $\mathbf{c}^{i, \mathcal{D}}_{\text{out}}$ after the discretization as follows:
\begin{equation}
\label{eqn:dis_com1}
\begin{aligned}
    \mathbf{c}^{i, \mathcal{D}}_{\text{in}}(t) &= \begin{bmatrix}
               \mathbf{c}^{i, \mathcal{D}}_{\text{in}, 1}(t)^T &
               \mathbf{c}^{i, \mathcal{D}}_{\text{in}, 2}(t)^T &
               \hdots &
               \mathbf{c}^{i, \mathcal{D}}_{\text{in}, P}(t)^T 
             \end{bmatrix}^T,\\
        \mathbf{c}^{i, \mathcal{D}}_{\text{in}, p}(t) &= \begin{cases} 
      \frac{1}{|\mathcal{P}^i_p(t)|} \sum\limits_{v_j \in \mathcal{P}^i_p(t)} \mathbf{c}_{j i}(t-1) &,   |\mathcal{P}^i_p(t)| \neq 0,  \\
      \mathbf{0} &,   |\mathcal{P}^i_p(t)| = 0,
   \end{cases}
\end{aligned}
\end{equation}

\begin{equation}
\label{eqn:dis_com2}
\begin{aligned}
    \mathbf{c}^{i, \mathcal{D}}_{\text{out}}(t) &= \begin{bmatrix}
               \mathbf{c}^{i, \mathcal{D}}_{\text{out}, 1}(t)^T &
               \mathbf{c}^{i, \mathcal{D}}_{\text{out}, 2}(t)^T &
               \hdots &
               \mathbf{c}^{i, \mathcal{D}}_{\text{out}, P}(t)^T
    \end{bmatrix}^T,\\
\mathbf{c}_{i  j}(t) &= \mathbf{c}^{i, \mathcal{D}}_{\text{out}, p}(t), \forall v_j \in \mathcal{P}^i_p(t),
\end{aligned}
\end{equation}
with $p \in \{1, 2, \cdots, P\}$, where $\mathbf{c}^{i, \mathcal{D}}_{\text{in}}(t)$ and $\mathbf{c}^{i, \mathcal{D}}_{\text{out}}(t)$ are the concatenations of $P$ communication vectors after discretization, and $|\cdot|$ represents the cardinality  of a set.
This renders the dimensions of the communication inflow and outflow constant. Thus, the distributed policy we aim to learn is transformed into
\begin{equation}
    \pi_d(o^i_{\text{local}}(t), \mathbf{c}^{i, \mathcal{D}}_{\text{in}}(t)) = (Q_{i}(t), \mathbf{c}^{i, \mathcal{D}}_{\text{out}}(t)), \forall v_i \in V.
\end{equation} 
In this work, we assume that the neighbour discretization is pre-designed and task-specific. Learning the discretization rule is left for future work.

%To demonstrate the intuition behind the neighbour discretization, we give an example where an agent can observe the relative positions of all its neighbours. One possible discretization rule can be that all the neighbours that are closer than $r$ to the agent belong to group 1 and all the other agents belong to group 2. For this discretization rule, we are assuming the communication 

% \begin{itemize}
%     \item After the discretization, the communication inflow is averaged for each group of agents and the communication to the agents in the same group  are identical.
%     \item This allows the size of the communication input and output to be invarient w.r.t number of agents, thus enables scalable communication protocols learning
%     \item In this work, only pre-designed neighbour grouping rule is explored. Learning of neighbour discretization rule is left as future work.
% \end{itemize}

\subsection{Learning from a Centralized Policy}
Using supervised learning, our approach builds upon a pre-designed centralized policy $\pi_c$, which defines the action probability distributions of all agents based on global full observations. This can be formulated as follows:
\begin{align}
    \pi_c(o_{\text{global}}(t)) &= \{Q^*_{i}(t) | v_i \in V\},
\end{align}
where
\begin{itemize}
    \item $o_{\text{global}}$ is the global observation, which includes all the local observations $\{o^i_{\text{local}}|v_i \in V\}$ in the global frame.
    \item $Q^*_{i}$ is the action probability distribution of agent $v_i$ suggested by the centralized policy $\pi_c$.
\end{itemize}
The objective is to make the learned distributed policy $\pi_d$ behave as similar to the centralized policy $\pi_c$ as possible. 

Directly learning the communication protocols among the agents from the centralized policy is difficult because the centralized policy does not provide correct and labelled communication protocols among the agents. Based on the problem setup, the communication vectors sent at the previous time step are received at the current time step, resulting in a communication information flow across multiple time steps. To learn such a flow of information, we require the learning process to operate across multiple time steps as well. 

To formulate the learning target, we first assume $\pi_d$ can be parametrized by $\bm{\theta}$ as $\pi_{d,\bm{\theta}}$. Over the course of performing task $T$ with $L$ time steps using a distributed policy $\pi_{d,\bm{\theta}}$, we can define the overall mapping we aim to achieve as
\begin{equation}
\begin{aligned}
     &\{o^i_{\text{local}}(t)| \forall v_i \in V,  t \in \{1, 2, \cdots, L\}\} \rightarrow \\ &\qquad\qquad\{a_i(t) | \forall v_i \in V,  t \in \{1, 2, \cdots, L\}\}.
\end{aligned}
\end{equation}
This mapping is modelled by a neural network consisting of identical components connected by communication vectors among the agents across multiple time steps (see Fig.~\ref{fig:giant_nn} for an example of the neural network). We recall that this neural network is a multi-step, multi-agent neural network (MSMANN), where the identical components are the distributed policy network for every agent at each time step. %The differentiable communication channels among the agents are the hidden states of the MSMANN. 
The backpropagation can be performed throughout the MSMANN to optimize the distributed policy with respect to a learning objective, which is minimizing $J(\bm{\theta})$:
\begin{equation}
    J(\bm{\theta}) = \frac{1}{KL} \sum_{v_i \in V\atop t \in \{1, 2, \cdots, L\}}^{} \mathcal{L}(Q_i(t), Q^*_i(t)),
\end{equation}
where $K$ is the number of agents and $\mathcal{L}(Q, Q^*)$ is the loss function for evaluating the difference between two action probability distributions. 

By minimizing $J(\bm{\theta})$, we can learn the distributed policy $\pi_d$ such that it behaves similarly to the centralized policy. With the guidance of the centralized policy, we hypothesize that this supervised learning process is more efficient than the reward-based learning, which has no direct guidance. 
An interesting possibility is to use reward-based learning for improvement after having completed supervised learning from the centralized policy. This possibility is left as future work.

\begin{figure}[t]
    \centering
    \includegraphics[width=0.4\textwidth]{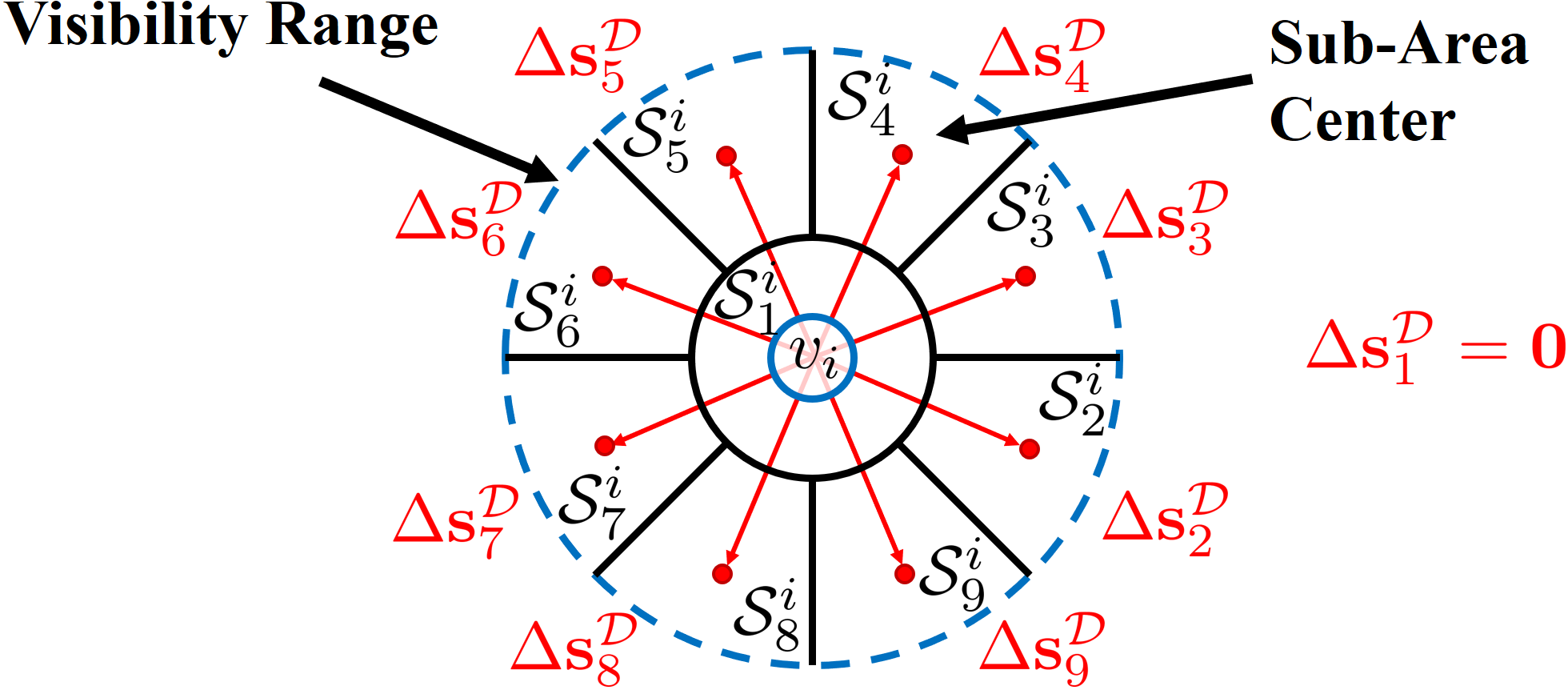}
    \caption{Radial discretization used in the rendezvous task. The visible area is discretized into 9 components. The vectors $\{\Delta\mathbf{s}^\mathcal{D}_1, \Delta\mathbf{s}^\mathcal{D}_2, \cdots \Delta\mathbf{s}^\mathcal{D}_9\}$ from the agent to the center of each area represent the discretized action space.}
    \label{fig:rad_dis}
    \vspace{-1em}
\end{figure}

\section{Simulation Setup: Two Different Tasks}
\label{sec:simulation_setup}
In order to demonstrate our proposed learning approach, we consider two coordination tasks: \textit{(i)} the rendezvous problem with limited vision range and \textit{(ii)} the particle assignment task. The former is well-studied in the control literature and our solution is compared against the state-of-the-art distributed control law; the latter is a task that has not been solved in the framework of distributed control before. 
%We consider two different distributed control tasks to evaluate our learning approach, Rendezvous with Limited visibility and Particle Assignment, which are introduced separately in the following sections.
\subsection{Rendezvous with Limited Visibility}
\label{sec:rendezvous_task}
We demonstrate our learning approach on the rendezvous task with limited visibility given \textit{(i)} its simplicity, and \textit{(ii)} its assumption of local interactions which fit our framework.

%This section is organized as follows. We first formulate the task that adapts to our learning approach and present some existing distributed policies for the task. Then, we define the neighbour discretization rule and the simplifications we make. Finally, we describe the centralized policy we design for our learning approach.%

%This section is organized as follows. We first formulate the task that adapts to our learning approach. Then, we define the neighbour discretization rule followed by the centralized policy designed for our learning approach. Finally, we introduce the existing distributed policies for comparison.

\subsubsection{Task Formulation}
Consider $K$ homogeneous agents located in a 2-dimensional plane with the position vectors $\{\mathbf{s}_1, \mathbf{s}_2, \cdots, \mathbf{s}_K\}$. Each agent is governed by double integrator dynamics and its position is controlled by a PD-type controller. The input to the controller is the desired position vector relative to the agent's position, which corresponds to the action $a$ of the agent. We define dynamic connectivity network $G(t) = (V, E(t))$ based on visibility:
%\begin{equation}
%\begin{aligned}
%    e(v_i, v_j) \in E(t) \iff \|\mathbf{s}_i(t) -  \mathbf{s}_j(t)\|_2 \leq d_{\text{lim}}
%\end{aligned}
%\end{equation}
\begin{equation}
    E(t) = \{ e(v_i, v_j) | v_i, v_j \in V, \|\mathbf{s}_i(t) -  \mathbf{s}_j(t)\|_2 \leq d_{\text{lim}} \},
\end{equation}
where $d_{\text{lim}}$ is the visibility range. 

We also define that each agent can only observe the relative positions of all its neighbours to itself as $o^i_{\text{local}}$.\footnote{The local frame, however, has the same heading as the global frame. This means that the local frame can be transformed to the global frame by only a translational transformation.} 
To formulate the action space and observation space, we discretize the 2-dimensional visibility space $\mathcal{S}^i$ of each agent into $P$ different components: $\{\mathcal{S}^i_1, \mathcal{S}^i_2, \cdots, \mathcal{S}^i_P\}$ (see Fig.~\ref{fig:rad_dis}). The observation of agent $v_i$ can be approximated by the number of agents in each discretized component: 
\begin{equation}
\begin{aligned}
    \label{eqn:dis_obs}
    \mathbf{o}^{i, \mathcal{D}}_{\text{local}}(t) &= 
    \begin{bmatrix}
              n^i_1(t) &
              n^i_2(t) &
              \hdots &
              n^i_P(t)
             \end{bmatrix}^T, \\
    \text{where } n^i_p(t) &= |\{v_j | \mathbf{s}_j \in \mathcal{S}^i_p, v_j \in N_i\}|,
\end{aligned}
\end{equation}
with $p \in \{ 1, 2, \cdots, P\}$. Using the same discretization, we restrict the action space to be the set of center points of the discretized components (see Fig.~\ref{fig:rad_dis}):
\begin{equation}
\label{eq:dis_act}
    \mathcal{A} = \{\Delta\mathbf{s}^\mathcal{D}_1, \Delta\mathbf{s}^\mathcal{D}_2, \cdots, \Delta\mathbf{s}^\mathcal{D}_P\}.
\end{equation}
The action probability distribution $Q_i(t)$ can be simplified into a vector of probabilities $\mathbf{q}_i(t)$ for choosing each discretized action:
\begin{equation}
    \label{eqn:vector_prob}
    \mathbf{q}_i(t) = \begin{bmatrix}
              Q_i(t, a = \Delta\mathbf{s}^\mathcal{D}_1) &
              %Q^i(t, a = \Delta\mathbf{s}^\mathcal{D}_2) \\
              \hdots &
              Q_i(t, a = \Delta\mathbf{s}^\mathcal{D}_P)
             \end{bmatrix}^T,
\end{equation}
where $Q_i(t, a = \Delta\mathbf{s}^\mathcal{D}_p)$ represents the probability of agent $v_i$ to choose action $\Delta\mathbf{s}^\mathcal{D}_p$ at time $t$ for all $p\in\{1,2,\cdots,P\}$. 

The objective of this task is to make all agents converge to a common location as quickly as possible. We assume that the dynamic connectivity network $G(t)$ is connected initially. The task performance is primarily evaluated based on the rendezvous time $t_{\text{RV}}$, which is defined as the smallest $t \in \{1, 2, \cdots, L\}$ that satisfies the following condition:
\begin{equation}
\label{eqn:rend_const}
    \|\mathbf{s}_i(t) - \mathbf{s}_j(t)\|_2 \leq \epsilon, \forall v_i, v_j \in V,
\end{equation}
where $\epsilon$ is a small constant that defines the maximum distance of the two farthest agents. If the constraint can never be satisfied, we classify this as a failure to converge ($t_{\text{RV}} = \infty$). After evaluating the task for multiple trials, we can define the convergence rate as follows: $CR\% = n_{\text{success}}/n_{\text{tot}} \times 100\%$, where $n_{\text{success}}$ is the number of successful trials and $n_{\text{tot}}$ is the number of total trials.

\subsubsection{Neighbour Discretization}
The neighbour discretization rule builds upon the space discretization performed in the task formulation. Using the relative positions to the neighbouring agents, we can define the partition as follows:
\begin{equation}
\begin{aligned}
   \mathcal{P}^i_p = \{v_k | \mathbf{s}_k \in \mathcal{S}^i_p, \forall v_k \in N_i\}, 
\end{aligned}
\end{equation}
for all $p \in \{1, 2 \cdots, P\}, v_i \in V$, where $\mathbf{s}_k$ is the position vector of agent $v_k$ and $\mathcal{P}^i_p$ is the $p^\text{th}$ discretization group of agent $v_i$'s neighbours $N_i$ under the discretization rule $\mathcal{D}$.

Under this rule, all the neighbouring agents in the same discretized component belong to the same discretization group. The intuition behind this is that the neighbouring agents with similar relative positions should receive similar communication vectors, and the communication vectors they are sending back should also be similar.

\subsubsection{Centralized Policy}
The centralized policy we use computes an optimal rendezvous coordinate $\mathbf{s}_{\text{optimal}}$ that minimizes $t_{\text{RV}}$. It can be proved that $\mathbf{s}_{\text{optimal}}$ is the center of the smallest circle that encloses all the agents (the smallest enclosing circle). By moving towards the optimal rendezvous coordinate, the optimal solution can be achieved.
To adapt to the discrete action space $\mathcal{A}$, the optimal action $a^*_i(t)$ for each agent is the closest one to the optimal coordinate:
%\begin{equation}
%\label{eqn:choose_action}
%\begin{aligned}
%    \argmin a^*_i(t) = \Delta\mathbf{s}_p^\mathcal{D} \in \mathcal{A}, \text{ such that} \\
%    \|\mathbf{s}_{\text{optimal}}(t) - \mathbf{s}_i(t) - \Delta\mathbf{s}_p^\mathcal{D}\|_2 \text{ is minimized}.
%\end{aligned}
%\end{equation}
\begin{equation}
\label{eqn:choose_action}
\begin{aligned}
    a^*_i(t) = \argmin_{\Delta\mathbf{s}^\mathcal{D}_p \in \mathcal{A}} 
    \|\mathbf{s}_{\text{optimal}}(t) - \mathbf{s}_i(t) - \Delta\mathbf{s}^\mathcal{D}_p\|_2, 
\end{aligned}
\end{equation}
%$a^*_i(t) = \Delta\mathbf{s}_p^\mathcal{D} \in \mathcal{A}, \text{ such that }  \|\mathbf{s}_{\text{optimal}}(t) - \mathbf{s}_i(t) - \Delta\mathbf{s}_p^\mathcal{D}\|_2 \text{ is minimized}.$

Since this centralized policy is deterministic, we define the probability distribution $Q^*_i$ over $\mathcal{A}$ as follows: %by defining the probability of choosing each possible action:
\begin{equation}
    Q^*_i(t, a) = \begin{cases} 
      1 & a = a^*_i(t)  \\
      0 & a \neq a^*_i(t) 
   \end{cases}, \forall a \in \mathcal{A},
\end{equation}
where $Q^*_i(t, a)$ is the probability of $v_i$ choosing action $a$ at time $t$ suggested by the centralized policy $\pi_c$. This distribution can be represented by a vector $\mathbf{q}^*_i$ similar to Eq.~\ref{eqn:vector_prob}.%, which will produce an one-hot vector in which all probabilities are 0 except the probability for the optimal action.

\subsubsection{Existing Distributed Policies}
 To the best of our knowledge, the circumcenter law is the state-of-the-art distributed policy for this task that guarantees convergence on single integrator dynamics~\cite{francis2016flocking}. This result can be simply extended to second-order systems with double integrator dynamics, which we consider for this task. The circumcenter law is defined as follows: At each time step, all agents pursue the circumcenter of the point set consisting of its neighbours and of itself (or the center of the smallest enclosing circle)~\cite{francis2016flocking}. To adapt this control law to our control framework described in Section~\ref{sec:framework}, we choose the agent's action $a_i$ based on the relative position of the agent to the circumcenter, equivalently to Eq.~\ref{eqn:choose_action}. The communication inflow $c^i_{\text{in}}$ and outflow $c^i_{\text{out}}$ consist only of zero vectors at all time steps. %The observation $o^i_{local}$ is defined as the relative position of its neighbours to itself.

There are other existing distributed policies such as the averaging law and cyclic pursuit. The averaging law requires all agents to pursue the average coordinate of its neighbours and of itself. However, it suffers from convergence issues~\cite{bullo2009distributed}. Cyclic pursuit usually assumes a fixed connectivity network, which is not given in this task~\cite{francis2016flocking}.

\subsection{Particle Assignment}
We also propose a new distributed robotic task that has no existing solution, to the best of our knowledge. The objective is to move all agents to target points such that all target points are covered by an agent using agents' local observations of the neighbouring agents and target points, and communication. Different from the cooperative navigation task defined in \cite{lowe2017multi} that assumes unlimited visibility range, we assume limited visibility range.
\subsubsection{Task Formulation}
We consider $K$ homogeneous agents with position vectors $\{\mathbf{s}_1, \mathbf{s}_2, \cdots, \mathbf{s}_K\}$ in a 2-dimensional plane. The agent dynamics, PD-controller, discrete action space $\mathcal{A}$, connectivity network $G(t)$ and discretization rule $\mathcal{D}$ are the same as described in the rendezvous task. Adding on top of these, we introduce $K$ target points $\bar{V} = \{\bar{v}_1,\bar{v}_2, \cdots, \bar{v}_K\}$ with the position vectors $\{\bar{\mathbf{s}}_1, \bar{\mathbf{s}}_2, \cdots, \bar{\mathbf{s}}_K\}$. We can define $\bar{G}(t) = (\bar{V}, \bar{E}(t))$ as the connectivity of the target points based on the visibility: %with $\bar{E}(t) = \{ e(\bar{v}_i, \bar{v}_j) | \bar{v}_i, \bar{v}_j \in \bar{V}, \|\bar{\mathbf{s}}_i(t) -  \bar{\mathbf{s}}_j(t)\|_2 \leq d_{\text{lim}}  \}$.
\begin{equation}
    \bar{E}(t) = \{ e(\bar{v}_i, \bar{v}_j) | \bar{v}_i, \bar{v}_j \in \bar{V}, \|\bar{\mathbf{s}}_i(t) -  \bar{\mathbf{s}}_j(t)\|_2 \leq d_{\text{lim}} \},
\end{equation}
where $d_{\text{lim}}$ is the limited visibility range.
% \begin{equation}
% \begin{aligned}
%     e(\bar{v}_i, \bar{v}_j) \in \bar{G}(t) \iff \|\bar{\mathbf{s}}_i(t) -  \bar{\mathbf{s}}_j(t)\|_2 \leq d_{\text{lim}}, \\
%     \forall \bar{v}_i, \bar{v}_j \in \bar{V}.
% \end{aligned}
% \end{equation}

We define the set of covered target points $\bar{V}_{\text{cov}}$ as follows:
\begin{equation}
    \bar{V}_{\text{cov}}(t) = \{\bar{v}_i | \bar{v}_i \in \bar{V}, \exists v_j \in V,  \|\mathbf{s}_j(t) - \bar{\mathbf{s}}_i(t)\|_2 < \epsilon \},
\end{equation}
where $\epsilon$ is a small constant that defines the distance requirement for covering a target point.

We assume that $G(t)$, $\bar{G}(t)$ are both connected initially, and at least one agent can observe one target point initially. Each agent also has a pre-defined potential field layer as a collision avoidance mechanism on top of the PD-controller.
For convenience, we can define the neighbouring target points as follows:
\begin{equation}
    \bar{N}_i(t) = \{\bar{v}_j | \bar{v}_j \in \bar{V}, \|\bar{\mathbf{s}}_j(t) -  \mathbf{s}_i(t)\|_2 \leq d_{\text{lim}}\}.
\end{equation}
The observation of each agent is defined as the relative positions of its neighbouring agents and target points to itself. Using the same space discretization in Section  \ref{sec:rendezvous_task}, we represent the observation as follows:
\begin{equation}
\label{eqn:dis_obs2}
\begin{aligned}
    \mathbf{o}^{i, \mathcal{D}}_{\text{local}}(t) &= 
    \left[\begin{matrix}
              n^i_1(t) &
              n^i_2(t) &
              \hdots &
              n^i_P(t) 
              \end{matrix} \right.\\
              &\qquad
              \left.\begin{matrix}
              \bar{n}^{i}_{1,\text{cov}}(t) &
              \bar{n}^{i}_{2,\text{cov}}(t) &
              \hdots &
              \bar{n}^{i}_{P,\text{cov}}(t) 
             \end{matrix} \right.\\
              &\qquad
              \left.\begin{matrix}
              \bar{n}^{i}_{1,\text{uncov}}(t) &
              \bar{n}^{i}_{2,\text{uncov}}(t) &
              \hdots &
              \bar{n}^{i}_{P,\text{uncov}}(t) 
             \end{matrix}\right]^T, \\
    \text{where } n^i_p &= |\{v_j | \mathbf{s}_j \in \mathcal{S}^i_p, v_j \in N_i\}|, \\ \bar{n}^{i}_{p, \text{cov}} &= |\{\bar{v}_j | \bar{\mathbf{s}}_j \in \mathcal{S}^i_p, \bar{v}_j \in \bar{N}_i, \bar{v}_j \in \bar{V}_{\text{cov}}\}|, \\ \bar{n}^{i}_{p, \text{uncov}} &= |\{\bar{v}_j | \bar{\mathbf{s}}_j \in \mathcal{S}^i_p, \bar{v}_j \in \bar{N}_i, \bar{v}_j \not\in \bar{V}_{\text{cov}}\}|.
\end{aligned}
\end{equation}
To evaluate the performance, we define $t_{\text{PA}}$ as the smallest $t \in \{1, 2, \cdots, L\}$ such that  
%\begin{equation}
%\label{eqn:complete_cons}
%     \bar{V}_{\text{cov}}(t) = \bar{V}%\exists v_j \in V \text{ such that } \|\mathbf{s}_j - \mathbf{s}^*_i\|_2 < \epsilon
%\end{equation}
$\bar{V}_{\text{cov}}(t) = \bar{V}$.
If the constraint can never be satisfied, $t_{\text{PA}} = \infty$. The convergence rate $CR\%$ is defined as in the rendezvous task. 

\subsubsection{Centralized Policy}
We design our centralized policy to be an optimal assignment of agents to target points so that the maximum distance travelled by any agent is minimized. This leads to the minimization of the completion time $t_{\text{PA}}$. The optimization is done by using the Hungarian algorithm~\cite{kuhn1955hungarian}.

\section{Simulation Results}
\label{sec:simulation_result}
In this section, we demonstrate the training details and our simulation results for the two distributed robotic tasks. %We evaluated our approach with metric, task completion time.  %For these simulation results, we remove all the examples where the convergence can not be achieved because including these results with the infinity completion time would blow up the average performance, which is not desired for our comparison purpose.%
\subsection{Training Details}
To perform this learning task, we use a DNN as the distributed policy network to model the mapping we aim to learn:
\begin{equation}
    (\mathbf{o}^\mathcal{D}_{\text{local}}, \mathbf{c}^{\mathcal{D}}_{\text{in}}) \rightarrow (\mathbf{q}, \mathbf{c}^{\mathcal{D}}_{\text{out}}).
\end{equation}
where $\mathbf{o}^\mathcal{D}_{\text{local}}$ is the local observation after discretization (see Eq.~\ref{eqn:dis_obs} for the rendezvous task, Eq.~\ref{eqn:dis_obs2} for the particle assignment task), $\mathbf{q}$ is the action probability vector after discretization (see Eq.~\ref{eqn:vector_prob}), $\mathbf{c}^{\mathcal{D}}_{\text{in}}$ and $\mathbf{c}^{\mathcal{D}}_{\text{out}}$ are the communication vectors after discretization (see Eq.~\ref{eqn:dis_com1} and \ref{eqn:dis_com2}).

For both tasks, the DNN is a feedforward neural network (2 layers with 32 neurons per layer for the rendezvous task; and 4 layers with 128 neurons per layer for the particle assignment task). The probability distribution $\mathbf{q}$ is obtained by a softmax layer. The loss function between the two probability distributions $\mathbf{q}$ and $\mathbf{q^*}$ is the cross entropy. To train the DNN, we run $n_{\text{batch}}$ simulations in parallel with different initial setups. %During the simulations, we use the neural network prediction to provide the action distribution for each agent. The action performed by each agent is directly sampled from this distribution provided. 
During the simulations, each agent's action is sampled from the action distribution predicted by the distributed policy network. After every $\ell$ time steps, we construct the MSMANN model (using the DNN as the identical components) based on the dynamic connectivity of the agents in the past $\ell$ time steps. Using the action probability distribution from the centralized policy, we perform the Adam algorithm through backpropagation on this MSMANN to minimize the average of loss functions over $n_{\text{batch}}$ parallel simulations~\cite{kingma2014adam}. Each cost function is an approximation of the overall learning objective:
\begin{equation}
\label{eqn:truncated_cost}
    J_{\text{truncated}}(\bm{\theta}) = \frac{1}{K\ell} \hspace{-0em} \sum_{\substack{v_i \in V \\ t \in \{t_c - \ell + 1, \cdots, t_c\}}}^{}\hspace{-2em} \mathcal{L}(Q_i(t), Q^*_i(t)),%\footnote{$\{t_c - \ell + 1:t_c\}$ is the short form for $\{t_c - \ell + 1, t_c - \ell + 2, \cdots, t_c\}$}
\end{equation}
where $t_c$ is the current time step and $\ell$ is the number of time steps we back-propagate. In this work, we choose $n_{\text{batch}} = 10$ for both tasks and $\ell = 6$ for the rendezvous task, and $\ell = 3$ for the particle assignment task. For the rendezvous task, we further simplify the communication inflow as the sum of the $P$ communication vectors after discretization ($\{\mathbf{c}^{i,\mathcal{D}}_{\text{in}, 1}, \cdots, \mathbf{c}^{i,\mathcal{D}}_{\text{in}, P}\}$) instead of the concatenation of these $P$ communication vectors as in Eq.~\ref{eqn:dis_com2}. This allows us to better analyze the communication content learned. %in Section~\ref{sec:com_analysis}.
\begin{figure}[t]
    \centering
    \includegraphics[width=8.5cm]{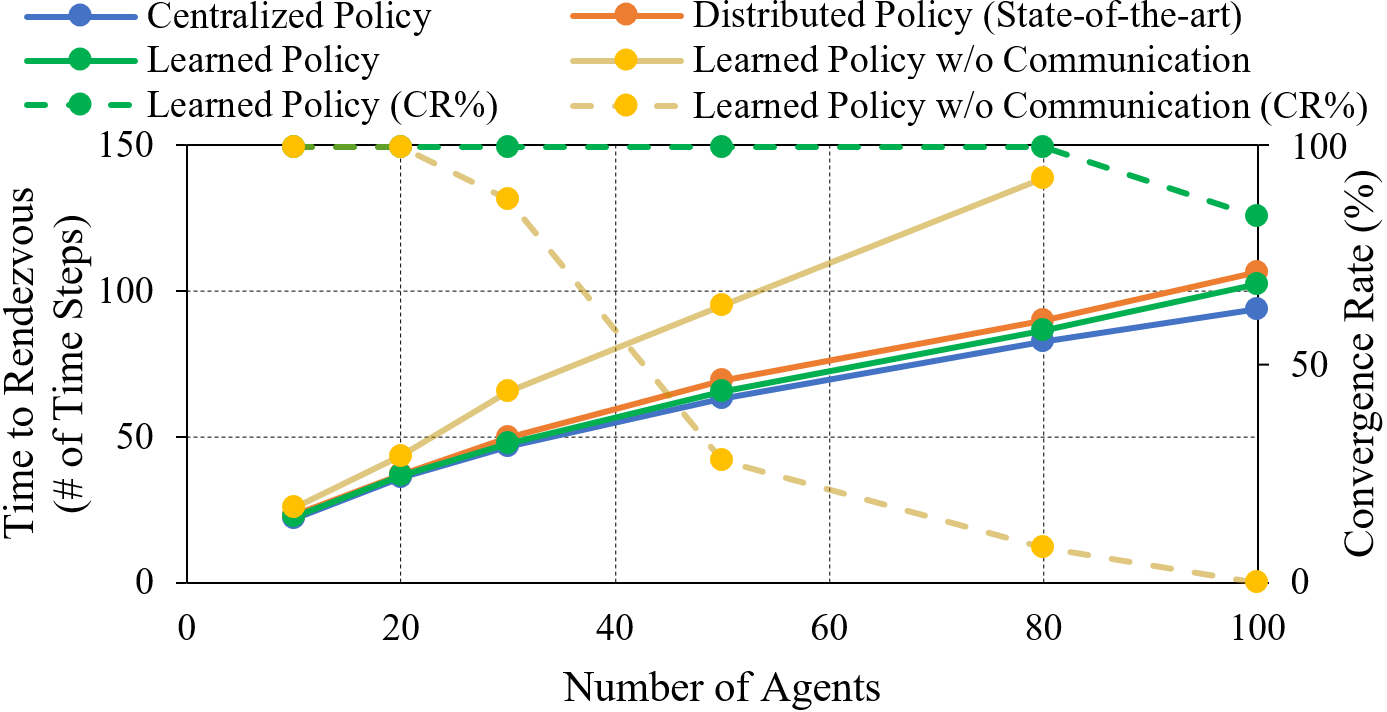}
    \caption{Performance comparison of centralized policy, state-of-the-art distributed policy, learned policy with communication enabled and learned policy without communication on the rendezvous task with different number of agents. Each data point is the average of 25 trials with random initial conditions. We explicitly exclude the examples that fail to converge ($t_{\text{RV}} = \infty$) since including these examples would blow up the average performance, which is not desired for our comparison purposes. Instead, we show the convergence rate. A video of the simulations is available online at: \url{http://tiny.cc/DNNswarm}.}
    \label{fig:res1}
    \vspace{-1em}
\end{figure}
\subsection{Rendezvous with Limited Visibility}
\label{sec:rendezvous_res}
For the rendezvous task, we train our distributed policy network on 10 agents with random initial positions and limit the size of the communication vector $\mathbf{c}_{ij}$ to be $n=25$. We evaluate the performance of the DNN on scenarios with different number of agents where the agent density is similar to the training cases. Agent density is defined as the ratio of number of agents to the area of the smallest circle that encloses all the agents. 

We compare the performance of our learning approach against the state-of-the-art circumcenter distributed control law and the centralized policy described in Section~\ref{sec:rendezvous_task}. We also show the performance of our learning approach without communication (i.e., $n=0$).
\begin{figure}[t]
    \centering
    \includegraphics[width=0.45\textwidth]{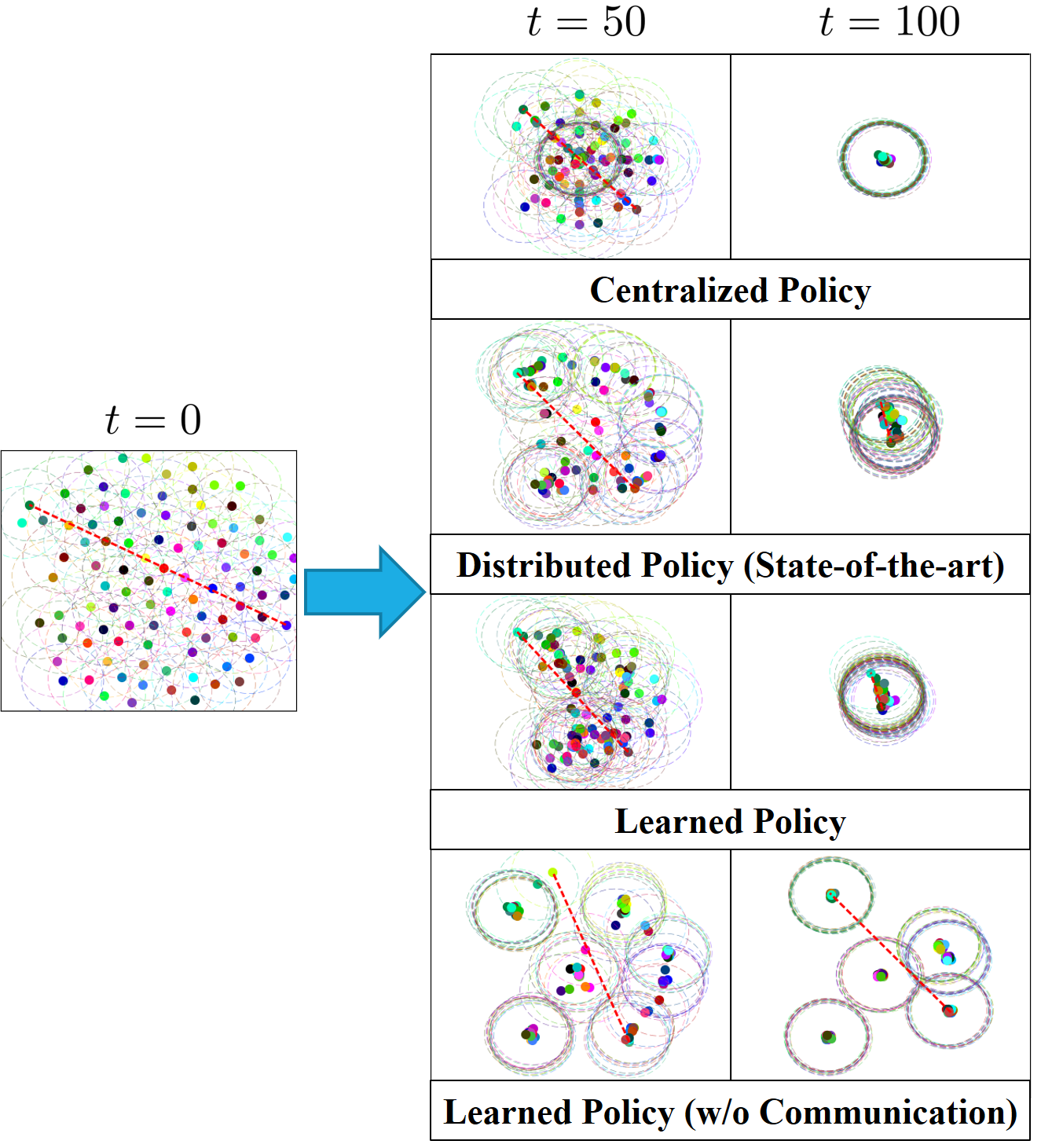}
    \caption{Comparison of the centralized policy, the state-of-the-art distributed policy, and the learned distributed policy with communication enabled and disabled on a 100-agent rendezvous task. The dots are the agents; the circles are the visible regions of the agents; the red line represents the distance of the two farthest agents. We show that the learned distributed policy can resemble the behaviour of the centralized policy significantly better when communication is available.}
    \label{fig:res2}
    \vspace{-0em}
\end{figure}
\begin{figure}[t]
    \centering
    \includegraphics[width=8.5cm]{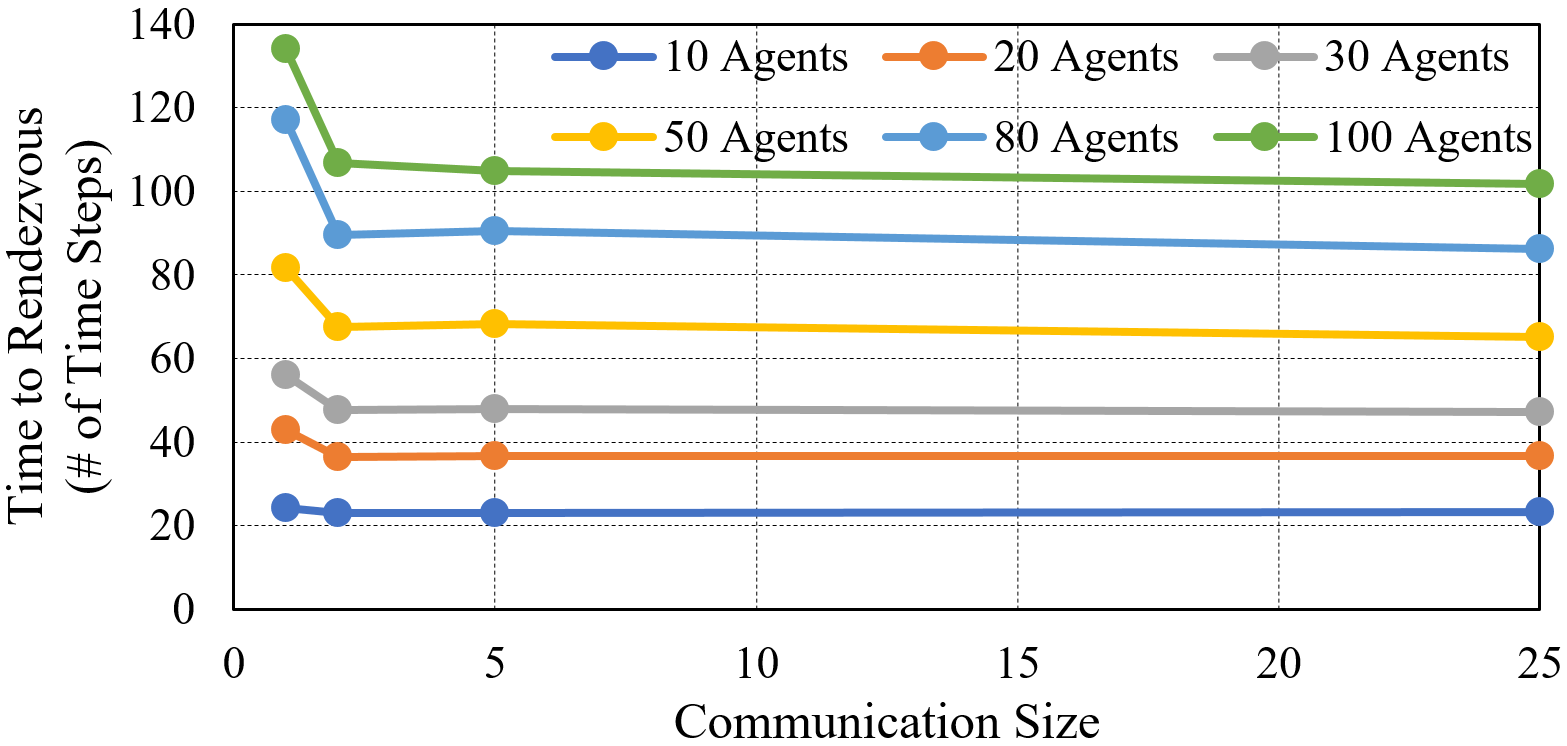}
    \caption{The performance of the learned policy for different communication vector sizes on the rendezvous task. Each data point is the average of 10 trials with different random initial conditions. A sudden drop in performance from $n = 2$ to $n = 1$ can be observed.}
    \label{fig:res3}
    \vspace{-1.5em}
\end{figure}
In Fig.~\ref{fig:res1}, we demonstrate that our learning approach consistently outperforms the state-of-the-art distributed policy for different numbers of agents. However, the learning approach without communication performs poorly under almost all circumstances, which demonstrates the necessity of inter-agent communications in resembling the behaviours of the centralized policy (see Fig.~\ref{fig:res2}). Note that the circumcenter control law does not require communication because it behaves qualitatively different than the centralized strategy.

%As another illustration, we include an example of rendezvous task with 100 agents following four different policies in Fig.~\ref{fig:res2}. From the convergence pattern, we can observe that our learning approach has successfully resemble some behaviours from the centralized policy, which is drastically different from the distributed circumcenter control law (state-of-the-art). It implies that the learned distributed policy would benefit from centralized policy and then outperform the existing distributed control law. From the last row, we also learn that communication is necessary for all agents to converge to the same location.

However, the convergence rate can drop significantly for scenarios with large number of agents, not included in the training data. %In Fig.~\ref{fig:res1b}, we show the successful rate of our learning approach drops as the number of agents increases. 
This is a generalization issue of the DNN learning as it might over-fit to the simple situations that the model is trained on.
%There are still limitations as the complexity of the larger-scale agent swarm tasks can grow far beyond the training situations in a smaller-scale.
Provided this limitation, our learning approach still demonstrates the ability to learn an effective distributed policy with reasonable scalability on this rendezvous task: we train with 10 agents and test on up to 100. 

% \begin{itemize}
%     \item Rendezvous with limited visibility (a comparison among four configurations: Centralized, State-of-the-art Distributed, Learned w/ Communication, Learned w/o Communication).
%     \item The approach successfully learns the behavior of centralized policy and improves the state-of-the-art distributed policy with extra communications among the agents
%     \item With no communication, the behavior cannot be learned properly
%     \item The approach demonstrates some scalability on up to 100 agents but show some limitation on lager scale. 
% \end{itemize}

\subsection{Analysis of Communication Learned}
\label{sec:com_analysis} 
\begin{figure}[t]
    \centering
    \includegraphics[width=8.5cm]{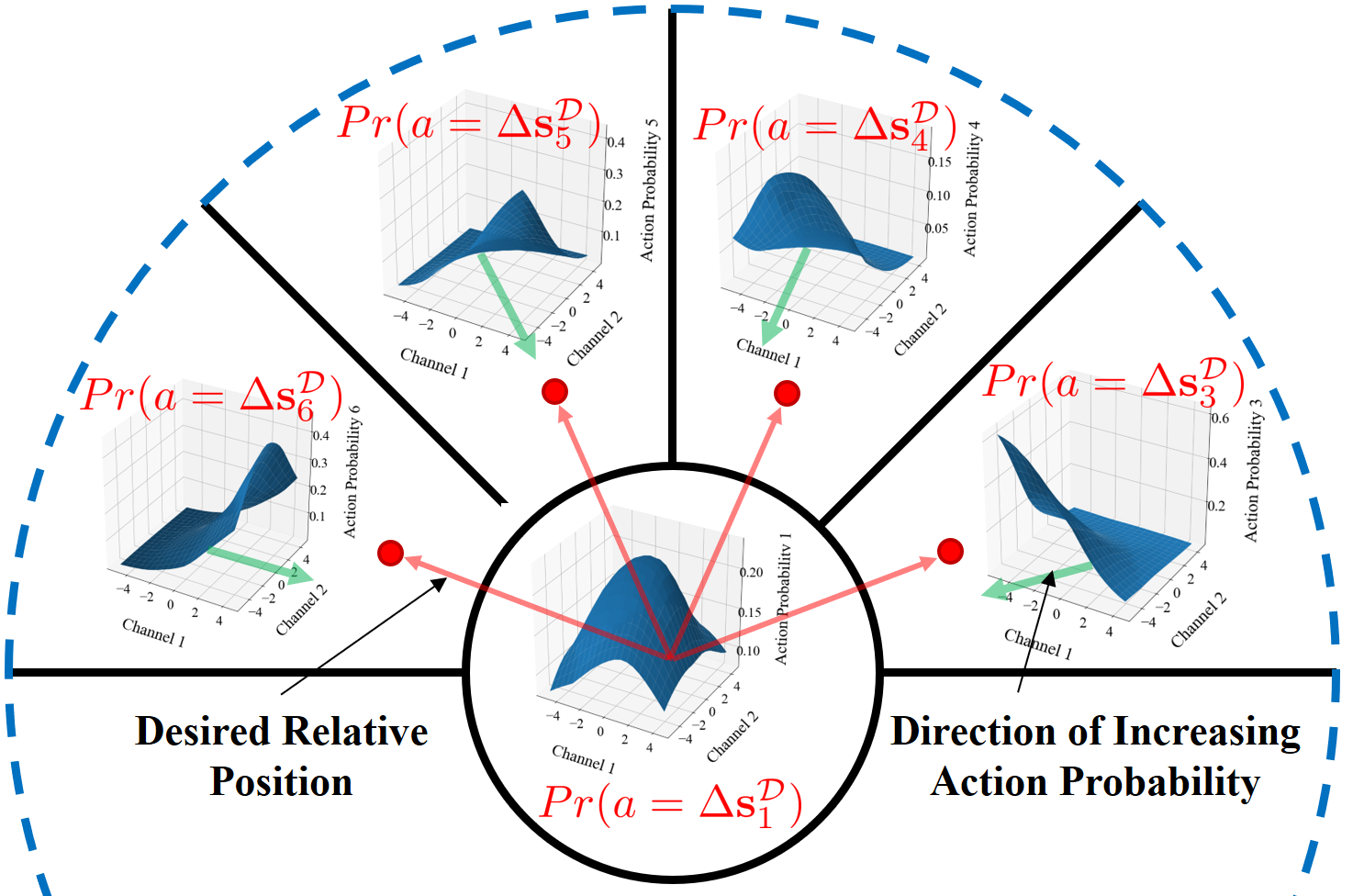}
    \caption{The probability of choosing each discretized action ($q_1, q_2, \cdots, q_9$) given the communication inflow values. We show five of them in this figure. ``Channel 1" represents $c_1$ and ``Channel 2" represents $c_2$. It can be observed that in each discretized area, the direction of the increasing action probability is always opposite to the desired relative position of the corresponding action in the discretized area.}
    \label{fig:com_analysis}
    \vspace{-1em}
\end{figure}

We demonstrate that reducing the size of the communication vectors leads to a decrease in task performance with a significant drop from $n=2$ to $n=1$ (see Fig.~\ref{fig:res3}). To provide more insights into this result, we choose to analyze the learned distributed policy with communication size $n = 2$ because it is relatively easy to visualize while achieving comparable performance. We keep the observation input of the model $\mathbf{c_{\text{in}}^{\mathcal{D}}}$ constant and observe the change in the model output $\mathbf{q}$ with the changing communication input. For the constant observation input, we assume a hypothetical situation where an agent has two neighbours that are located in exactly the opposite direction. For convenience, we define
%\begin{align}
%    \mathbf{c}^{\mathcal{D}}_{\text{in}}(t) &= \begin{bmatrix}
%              c_1 &
%              c_2
%             \end{bmatrix}^T, \\
%    \mathbf{q}(t) &= \begin{bmatrix}
%              q_1 &
%              q_2 &
%              \hdots &
%              q_P
%             \end{bmatrix}^T. %\\
    % \mathbf{c}^{\mathcal{D}}_{\text{out}}(t) &= \begin{bmatrix}
    %         %   \mathbf{c}^{i, \mathcal{D}}_{out, 1}(t) \\
    %         %   \mathbf{c}^{i, \mathcal{D}}_{out, 2}(t) \\
    %         %   \vdots \\
    %         %   \mathbf{c}^{i, \mathcal{D}}_{out, P}(t)
    %         %   \end{bmatrix} &= \begin{bmatrix}
    %           g_{1,1} &
    %           g_{1,2} &
    %           g_{2,1} &
    %           \hdots  &
    %           g_{P,1} &
    %           g_{P,2} \end{bmatrix}^T.
%\end{align}
$\mathbf{c}^{\mathcal{D}}_{\text{in}}(t) = \begin{bmatrix}
              c_1 &
              c_2
             \end{bmatrix}^T
$ and $\mathbf{q}(t) = \begin{bmatrix}
              q_1 &
             q_2 &
              \hdots &
              q_P
             \end{bmatrix}^T$.
%We aim to observe the change in the variables $q_1, q_2, \cdots, q_P, g_{1,1}, g_{1, 2}, \cdots, g_{P, 1}, g_{P,2}$ with respect to $f_1$ and $f_2$.
%we demonstrate the change in probability of each discretized action $q_1, q_2, \cdots, q_9$ while changing the values of the communication inputtu $f_1$ and $f_2$. 
In Fig.~\ref{fig:com_analysis}, we hypothesize that the two communication inflow values can be transformed into a 2-dimensional vector that is correlated to the tendency of choosing the action that is closest to the vector's direction. We can interpret the communication vector as an ``intent vector", which influences the tendency of the moving direction of the agent that receives this ``intent vector". This explains the sudden drop observed in performance from $n = 2$ to $n = 1$ since a one-dimensional communication vector cannot fully represent a 2-dimensional direction vector. 
\begin{figure}[t]
    \centering
    \includegraphics[width=8.5cm]{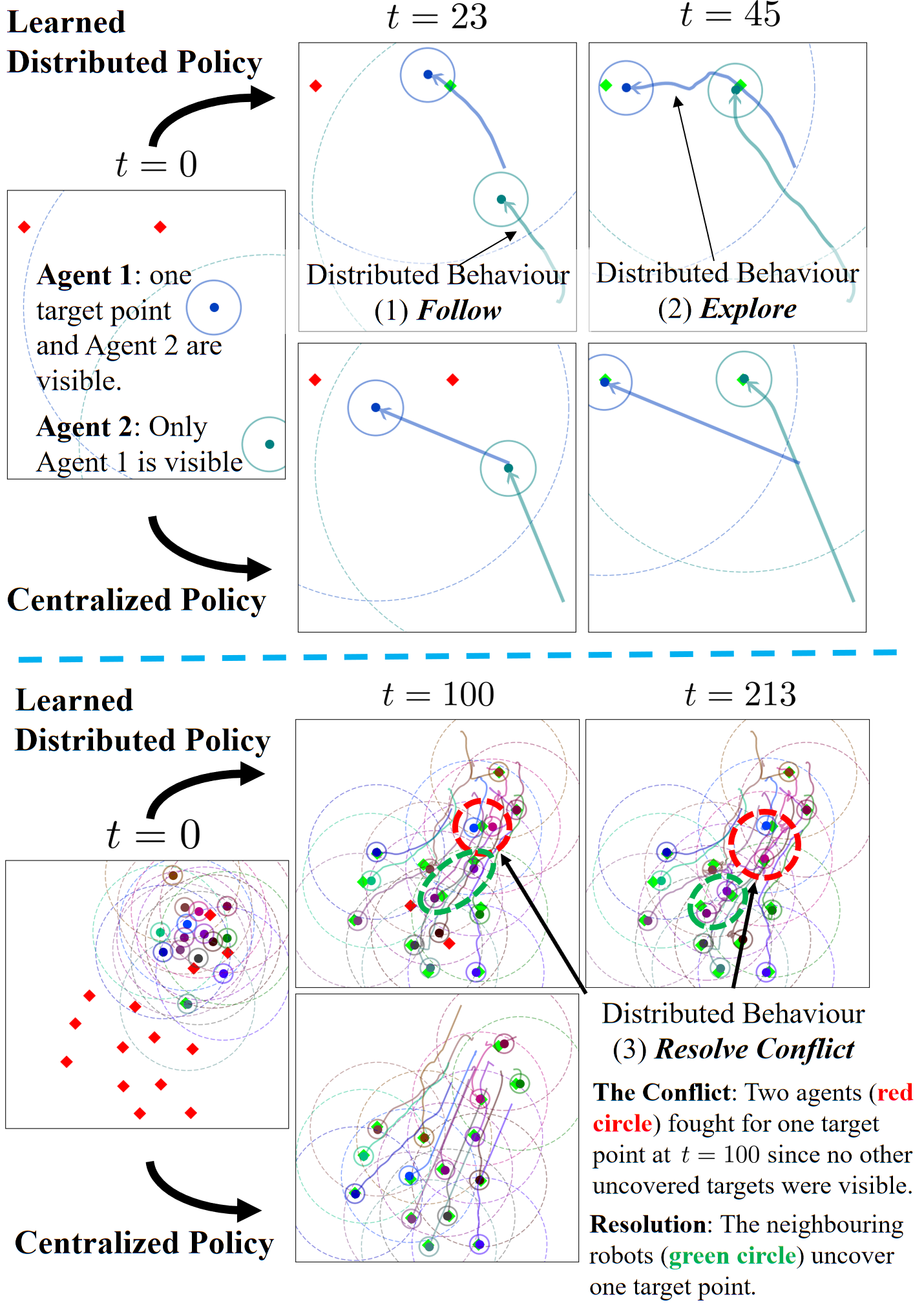}
    \caption{A behaviour comparison of the centralized policy and the learned distributed policy for 2-agent (top) and 15-agent (bottom) scenarios. Dots represent the agents; lines represent the trajectories of the agents; small circles represent the cover range $\epsilon$ of the agents; large dashed circles represent the visibility of the agents; and diamonds represent the target points. We demonstrate that agents controlled by our learned distributed policy are able to (1) follow other agents who see the targets when there is no target in its sight, (2) explore neighbouring targets rather than stopping at the nearest target, and (3) resolve target assignment conflicts. These distributed behaviours emerge from the learning of a centralized policy.}
    \label{fig:PA_example}
    \vspace{-1.5em}
\end{figure}
\begin{figure}[t]
    \centering
    \includegraphics[width=8.5cm]{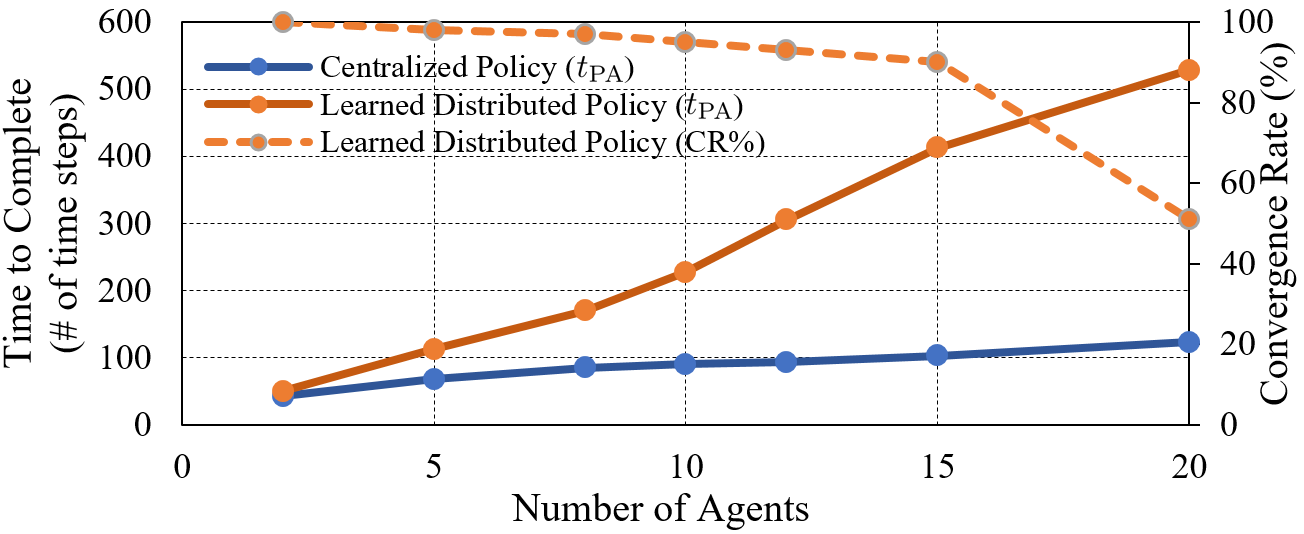}
    \vspace{-1em}
    \caption{A performance comparison of the centralized policy and the learned distributed policy on different numbers of agents for the particle assignment task. Each data point is the average of 100 trials. A video of the simulations is available online at: \url{http://tiny.cc/DNNswarm}}
    \label{fig:pa_performance}
    \vspace{-1em}
\end{figure}

\subsection{Particle Assignment}
For the particle assignment task, we also train our DNN on 10-agent scenarios and test our approach on various numbers of agents up to 20. Fig.~\ref{fig:PA_example} demonstrates some examples of the performance achieved with 2 and 15 agents and the emergence of distributed behaviours. In Fig.~\ref{fig:pa_performance}, we also show that the average performance of our learned distributed policy is comparable with the centralized policy when there are fewer agents. This approach suffers from convergence issues for larger swarms. %with a large of number of agents as compared against the rendezvous task. %
We hypothesize that this could be attributed to the inherent complexity of the particle assignment task. There are multiple aspects of the task that must be achieved: exploring, resolving assignment conflict, and staying connected with other agents. Achieving all aspects at once can be much more challenging as the number of agents increases.

\section{Conclusions and Future Work}
We present a DNN-based approach that learns distributed action and communication policies from well-designed centralized policies for homogeneous, distributed robotic system. %It is demonstrated that our learning approach can consistently outperform the state-of-the-art distributed policy for rendezvous task with limited visibility. %on different number of agents. 
%For the particle assignment distributed robotic task we define, the approach can achieve comparable performances to centralized policies, with the emergence of distributed behaviours %(illustrated in Fig.~\ref{fig:PA_example}) 
%that are necessary for completing the task. 
The main advantages of our proposed approach are summarized: \textit{(i)} this approach can be applied to various distributed robotic tasks given pre-designed centralized policies are available; \textit{(ii)} it requires little human expertise for task-specific control law and communication protocol designs; and \textit{(iii)} this approach is computationally efficient compared to other reward-based learning approaches. Moreover, the learned communication protocols reveal that meaningful messages are conveyed, which could potentially inspire the coordination and communication designs for real-world distributed robotic systems. Future work will address some of the observed convergence issues in the more complex scenarios, which may be due to over-fitting. %As a potential result of over-fitting the model to the training scenarios, our approach can suffer from convergence issues in the more complex scenarios. Potential solutions to these issues can be the wiser selections of training scenarios and discretization rules for communication. We left them as future works. 

%bring us more insights to design the communication protocols for real-world distributed robotic system.

%There are limitations of our approach, namely the difficulty for the learned model to be generalized to more complex scenarios. %that the model learned can be difficult to be generalized to more complex scenarios. %For rendezvous task, we scale up the number of agents from 10 to 100 with the similar performance. For the particle assignment task, we can also scale it up from 10 to 15. 

\bibliographystyle{IEEEtran}
%\bibliography{IEEEabrv,main}
\bibliography{IEEEabrv,bibliography}
\end{document}